\documentclass{article}

\usepackage{microtype}
\usepackage{graphicx}
\usepackage{subfigure}
\usepackage{booktabs} % for professional tables

\usepackage{hyperref}

\usepackage[accepted]{MOSS_camera_ready}

\usepackage{amsmath}
\usepackage{amssymb}
\usepackage{mathtools}
\usepackage{amsthm}

\usepackage[capitalize,noabbrev]{cleveref}

\theoremstyle{plain}

\theoremstyle{definition}

\theoremstyle{remark}

\usepackage[textsize=tiny]{todonotes}

\icmltitlerunning{Review, Remask, Refine: Process-Guided Block Diffusion for Text Generation}

\begin{document}

\onecolumn
\icmltitle{Review, Remask, Refine (R3): Process-Guided Block Diffusion for Text Generation}



\icmlsetsymbol{equal}{*}

\begin{icmlauthorlist}
\icmlauthor{Nikita Mounier}{equal,yy}
\icmlauthor{Parsa Ideahpour}{equal,yyy}
\end{icmlauthorlist}

\icmlaffiliation{yy}{Jerome Fisher M\&T Program, University of Pennsylvania, Philadelphia, USA}
\icmlaffiliation{yyy}{Computer Information Sciences, University of Pennsylvania, Philadelphia, USA}

\icmlcorrespondingauthor{Nikita Mounier}{nmounier@seas.upenn.edu}
\icmlcorrespondingauthor{Parsa Ideahpour}{ididea@seas.upenn.edu}

\icmlkeywords{Machine Learning, ICML}

\vskip 0.3in


\printAffiliationsAndNotice{\icmlEqualContribution} % otherwise use the standard text.

\begin{abstract}
    A key challenge for iterative text generation is enabling models to efficiently identify and correct their own errors. We propose Review, Remask, Refine (R3), a relatively simple yet elegant framework that requires no additional model training and can be applied to any pre-trained masked text diffusion model (e.g., LLaDA or BD3-LM). In R3, a Process Reward Model (PRM) is utilized for the \textbf{Review} of intermediate generated blocks. The framework then translates these PRM scores into a \textbf{Remask} strategy: the lower a block's PRM score, indicating potential mistakes, the greater the proportion of tokens within that block are remasked. Finally, the model is compelled to \textbf{Refine} these targeted segments, focusing its efforts more intensively on specific sub-optimal parts of past generations, leading to improved final output.
\end{abstract}

\section{Introduction}
\label{introduction}

Recent advancements in generative modeling have seen the rise of masked diffusion models (MDMs) for text generation, offering a compelling alternative to traditional autoregressive approaches \cite{LLaDA}. These models operate by iteratively refining a sequence, often starting from a masked state and progressively unmasking or de-noising tokens to produce coherent text. This iterative, non-autoregressive (or partially autoregressive when using block diffusion techniques \cite{BDLM_block_diffusion}) nature presents unique opportunities for controlled and high-quality text synthesis, allowing for the potential to revisit and improve discrete parts of the sequence.

However, a significant challenge lies in effectively guiding this iterative refinement process. Specifically, enabling these models to efficiently identify and correct their own earlier, less optimal generations is essential for maximizing output quality. While general remasking capabilities allow models to revisit previously generated tokens \cite{ReMDM_paper}, the strategy determining \textit{what} to remask is paramount for efficiency and impact.

To address this, we leverage the capabilities of Process Reward Models (PRMs). Unlike outcome-based reward models that only evaluate the final generated output, PRMs provide fine-grained feedback on the intermediate steps or components of a generation process (\cite{processbench}, \cite{prmlessons}). This makes PRMs exceptionally well-suited for guiding iterative models, as they can offer nuanced assessments of partially generated sequences or specific text blocks.

In this paper, we introduce \textbf{Review, Remask, Refine (R3)}, a novel, relatively simple yet elegant framework that combines the strengths of masked text diffusion models with the targeted guidance of PRMs. R3 facilitates a cycle of self-correction within the generation process:
\begin{enumerate}
    \item \textbf{Review:} A Process Reward Model (PRM) is utilized for the \textbf{Review} of the quality of currently generated text blocks.
    \item \textbf{Remask:} Based on the PRM scores obtained during the review, R3 implements a remasking strategy. Critically, the lower a block's PRM score -- indicating potential issues or ``mistakes'' -- the greater the proportion of tokens \textit{within} that block are remasked.
    \item \textbf{Refine:} The masked diffusion model is then prompted to refine these remasked segments, thereby focusing its generative capacity on the areas identified as weakest by the PRM.
\end{enumerate}

A key advantage of the R3 framework is that it requires \textbf{no additional training or fine-tuning of the underlying masked diffusion model} and can be readily integrated with various existing pre-trained models (e.g., LLaDA or BD3-LM). By strategically directing the model to revisit and improve specific sub-optimal segments of its past generations, R3 offers an efficient pathway to higher-quality text. Our contributions are: (1) the R3 framework for PRM-guided iterative refinement in masked text diffusion; (2) the specific mechanism of PRM-score-dependent proportional remasking for targeted error correction; and (3) a demonstration of how this approach can enhance text generation without requiring model retraining.

\section{Related Work}
Our Review, Remask, Refine (R3) framework integrates concepts from masked diffusion models for text, process-guided supervision, and iterative refinement strategies.

\textbf{Masked and Block Diffusion Models} Recent text generation approaches include masked diffusion models like LLaDA \cite{LLaDA}, which iteratively denoise sequences, and Block Diffusion (BD3-LMs) \cite{BDLM_block_diffusion}, which combine autoregressive block-wise generation with intra-block discrete diffusion. These models offer iterative refinement capabilities. R3 is designed to leverage these models by intelligently guiding their block-wise refinement process.

\textbf{Process Reward Models (PRMs) for Guidance} Unlike Outcome Reward Models (ORMs) that score only final outputs \cite{ouyang2022training}, PRMs \cite{processbench, prmlessons} evaluate intermediate steps, providing granular feedback ideal for iterative generation. In R3, the ``Review'' stage employs a PRM, whose score \(S_b\) for a block \(b\) directly informs the subsequent ``Remask'' stage, focusing refinement on identified weaknesses.

\textbf{Iterative Refinement and Guided Remasking} The ability to correct prior generations is essential. \textbf{ReMDM} \cite{ReMDM_paper} allows already generated tokens in masked diffusion models to be remasked and updated at inference time. R3 advances this by introducing a PRM-guided remasking strategy. The proportion of tokens remasked within block \(b\), \(P_{\text{remask}}(b)\), is a decreasing function of its PRM score \(S_b\) (i.e., \(P_{\text{remask}}(b) = f(S_b)\) with \(f'(S_b) < 0\)). This targeted approach, requiring no retraining of the base model, distinguishes R3 by focusing refinement efforts on PRM-identified suboptimal segments, offering a simple yet effective integration of process supervision. Other strategies to enhance reasoning in dLLMs, such as the d1 framework \cite{d1}, adapt pre-trained masked dLLMs using a combination of supervised finetuning (SFT) and reinforcement learning (RL). This contrasts with R3's approach, which aims to improve generation quality at inference time without requiring additional post-training of the base text diffusion model.

\section{The R3 Framework}
We introduce the Review, Remask, Refine (R3) framework, an iterative inference-time procedure designed to enhance the quality of text generated by masked diffusion models. R3 leverages Process Reward Models (PRMs) to provide targeted feedback for error correction and refinement, without requiring retraining of the base generative model or the PRM. The framework operates by cyclically generating text in discrete blocks, assessing these blocks with a PRM, selectively remasking segments of blocks identified as suboptimal, and then employing the base diffusion model to refine these remasked portions.

\subsection{Core Components}
The R3 framework is built upon two essential pre-trained models. The base Masked Diffusion Model (\(M_{\text{diff}}\)) is a generative model proficient at infilling masked regions within a text sequence. Given an input sequence \(x\) containing masked tokens \(x_{\text{mask}}\), \(M_{\text{diff}}\) produces a prediction \(\hat{x}_0 = M_{\text{diff}}(x; x_{\text{mask}})\) for the original content. The Process Reward Model (\(M_{\text{PRM}}\)) evaluates the quality of individual text segments. For a given text block \(x_b\) conditioned on prior context \(C_b\), \(M_{\text{PRM}}\) outputs a scalar quality score \(S_b = M_{\text{PRM}}(x_b | C_b)\), where \(S_b \in [0, 1]\) and higher scores denote superior quality.

\subsection{The R3 Iterative Process with Windowed Refinement}
R3 constructs text block by block. A central feature of our approach is a windowed evaluation and batched refinement mechanism, allowing for efficient application of PRM guidance. The iterative process is as follows:

Let \(X^{(j)}\) be the sequence generated up to block \(j-1\).
For each new block \(j = 0, \dots, N_{\text{total}}-1\):
\begin{enumerate}
    \item \textbf{Initial Block Generation (Extend):}
    Generate the content for the current block \(x_j\) using \(M_{\text{diff}}\), conditioned on \(X^{(j)}\). This step involves an iterative demasking procedure where initially masked tokens for block \(x_j\) are progressively filled. The sequence is updated: \(X^{(j+1)} = X^{(j)} \oplus x_j\), where \(\oplus\) denotes concatenation.

    \item \textbf{Windowed Review (Score Blocks):}
    Periodically, typically every \(K\) blocks (the window size), the last \(K\) generated blocks \(W_j = \{x_{\max(0, j-K+1)}, \dots, x_j\}\) are evaluated. For each sequence in the batch, \(M_{\text{PRM}}\) assesses each block \(x_b \in W_j\), yielding a set of scores \(\mathcal{S}_{W_j} = \{S_{\max(0, j-K+1)}, \dots, S_j\}\). These scores are stored.

    \item \textbf{Refinement Trigger:}
    For each sequence in the batch, if its minimum score within the current window \(\min(\mathcal{S}_{W_j})\) falls below a predefined threshold \(\tau_{\text{thresh}}\), a refinement cycle is initiated for its window \(W_j\).

    \item \textbf{Candidate Generation for Refinement (Remask \& Propose):}
        For each sequence whose window \(W_j\) triggered refinement:
        \begin{itemize}
            \item \textbf{Calculate Remasking Probabilities:} For each block \(x_b \in W_j\) with score \(S_b\), a remasking probability \(P_R(S_b)\) is determined. \(P_R(S_b)\) is a monotonically decreasing function of \(S_b\); for instance, \(P_R(S_b) \propto \exp(\alpha_B (1-S_b))\), where \(\alpha_B\) is a scaling parameter. This ensures that lower-quality blocks are more likely to be remasked more extensively.
            \item \textbf{Token-level Remasking:} Within each block \(x_b \in W_j\), a proportion of its tokens, \(\rho_b = \beta_I \cdot \tilde{P}_R(S_b)\) (where \(\tilde{P}_R(S_b)\) might be a normalized or clipped version of \(P_R(S_b)\) and \(\beta_I\) is an intensity factor), are selected and replaced by `[MASK]' tokens. This results in a masked window \(W_{j, \text{masked}}\).
            \item \textbf{Propose Refinements:} \(N_S\) candidate refined versions of \(W_{j, \text{masked}}\), denoted \(\{\hat{W}_{j}^{(s)}\}_{s=1}^{N_S}\), are generated by applying \(M_{\text{diff}}\) to \(N_S\) copies of \(X^{(j-K+1)} \oplus W_{j, \text{masked}}\). This step is typically batched for efficiency.
        \end{itemize}

    \item \textbf{Candidate Scoring (Review Refined Candidates):}
    Each of the \(N_S\) refined candidate windows \(\hat{W}_{j}^{(s)}\) is scored using \(M_{\text{PRM}}\), yielding sets of scores \(\{\mathcal{S}_{W_j}^{(s)'}\}_{s=1}^{N_S}\).

    \item \textbf{Selection and Update (Refine):}
    For each sequence that underwent refinement, the optimal candidate window \(\hat{W}_{j}^{(s^*)}\) is selected from the \(N_S\) proposals. The selection is based on a metric \(\mathcal{M}(\cdot)\) applied to the candidate scores, e.g., \(s^* = \arg\max_s \mathcal{M}(\mathcal{S}_{W_j}^{(s)'})\), where \(\mathcal{M}\) could be the product of scores or the minimum score within the window. The main sequence \(X^{(j+1)}\) is updated by replacing its window \(W_j\) with the selected \(\hat{W}_{j}^{(s^*)}\), and the stored scores for these blocks are updated accordingly. If refinement was not triggered or no candidate offers improvement, the original blocks \(W_j\) are retained.
\end{enumerate}
This R3 cycle allows the model to progressively build text while continuously evaluating and correcting prior segments, thereby focusing refinement efforts on the passages most in need of improvement as identified by the PRM. The detailed pseudocode is presented in Algorithm~\ref{alg:r3_framework} below.

\subsection{Algorithm Pseudocode}
The R3 framework with windowed refinement is formalized in Algorithm \ref{alg:r3_framework}.

\begin{algorithm}[h!]
    \caption{R3 Framework with Windowed Refinement}
    \label{alg:r3_framework}
 \begin{algorithmic}[1]
    \STATE {\bfseries Input:} Initial prompt \(X^{(0)}\), Base Diffusion Model \(M_{\text{diff}}\), PRM \(M_{\text{PRM}}\)
    \STATE {\bfseries Parameters:} Total blocks \(N_{\text{total}}\), block length \(L_B\), window size \(K\), threshold \(\tau_{\text{thresh}}\), num samples \(N_S\), intensity \(\beta_I\), alpha \(\alpha_B\)
    \STATE Let \(X \leftarrow X^{(0)}\) ; \(S_{\text{all}} \leftarrow \emptyset\) (list of lists for scores per block per batch item)
    \FOR{block index \(j = 0\) {\bfseries to} \(N_{\text{total}}-1\)}
        \STATE \(x_j \leftarrow M_{\text{diff}}(X | \text{mask for block } j)\) \COMMENT{Generate current block}
        \STATE \(X \leftarrow X \oplus x_j\)
        \STATE Append placeholder for \(S_j\) to \(S_{\text{all}}\) for each batch item.
        \IF{ (\(j+1\)) \texttt{mod} \(K == 0\) {\bfseries or} \(j == N_{\text{total}}-1\) }
            \STATE Let window \(W = \{x_{\max(0, j-K+1)}, \dots, x_j\}\)
            \STATE For each item \(b_{\text{item}}\) in batch: \(S_{W}[b_{\text{item}}] \leftarrow \{M_{\text{PRM}}(x_b | X_{<b}) \text{ for } x_b \in W[b_{\text{item}}]\}\). Update \(S_{\text{all}}\).
            \FOR{each item \(b_{\text{item}}\) in batch}
                \IF{ \(\min(S_{W}[b_{\text{item}}]) < \tau_{\text{thresh}}\) }
                    \STATE For each \(x_b \in W[b_{\text{item}}]\), calculate remasking prob \(P_R(S_b)\) using \(\alpha_B\).
                    \STATE For \(s=1 \dots N_S\):
                        \STATE Create \(X^{(s)}_{\text{masked}}\) by masking tokens in \(W[b_{\text{item}}]\) of a copy of \(X[b_{\text{item}}]\) based on \(P_R(S_b)\) and \(\beta_I\).
                        \STATE \(\hat{W}^{(s)} \leftarrow \{M_{\text{diff}}(X^{(s)}_{\text{masked}} | \text{mask for block } b) \text{ for each block } b \text{ in window}\}\) \COMMENT{Refine candidates}
                    \STATE Score all \(\{\hat{W}^{(s)}\}_{s=1}^{N_S}\) using \(M_{\text{PRM}}\) to get \(\{\mathcal{S}_{W}^{(s)'}\}\).
                    \STATE \(s^* \leftarrow \arg\max_s \mathcal{M}(\mathcal{S}_{W}^{(s)'})\).
                    \STATE Replace window in \(X[b_{\text{item}}]\) with \(\hat{W}^{(s^*)}\).
                    \STATE Update corresponding scores in \(S_{\text{all}}[b_{\text{item}}]\) with \(\mathcal{S}_{W}^{(s*)'}\).
                \ENDIF
            \ENDFOR
        \ENDIF
    \ENDFOR
    \STATE {\bfseries Output:} Final sequence \(X\)
 \end{algorithmic}
 \end{algorithm}

\section{Experiments}
We evaluate the efficacy of our Review, Remask, Refine (R3) framework, focusing on its ability to correct errors in mathematical reasoning tasks and comparing its performance and computational characteristics against relevant baselines.

\subsection{Experimental Setup}
We utilize LLaDA-8B-Instruct \cite{LLaDA} as the base masked diffusion model (\(M_{\text{diff}}\)) for generation and refinement. For the ``Review" stage (and for the BoN baseline), we employ Qwen2.5-Math-PRM-7B \cite{prmlessons} as our Process Reward Model (\(M_{\text{PRM}}\)).

Experiments were conducted on a subset of 127 questions from the MATH 500 dataset, focusing on problems requiring step-by-step derivations.

Key hyperparameters for R3: sampling temperature \(0.8\), PRM threshold \(\tau_{\text{thresh}} = 0.8\), backmasking intensity \(\beta_I = 0.8\), candidate samples \(N_S = 5\), score mapping factor \(\alpha_B = 10.0\). Sequences used 16 blocks of 32 tokens (512 total) with 128 demasking steps. Window size \(K\) was varied across experiments.

We compare R3 against two main baselines. First, pass@1 represents standard sequential block-by-block generation from LLaDA-8B-Instruct with a sampling temperature of \(0.8\). No PRM guidance or re-evaluation is used. Second, Block-wise Best of N (B-BoN) generates \(N_S=5\) candidate versions for each block in the sequence using \(M_{\text{diff}}\). Each candidate block is then scored by \(M_{\text{PRM}}\), and the highest-scoring candidate block is selected to extend the sequence before proceeding to the next block. This represents a strong baseline that heavily utilizes the PRM at each step.

\subsection{Qualitative Example: Error Correction Cycle}
The R3 framework identifies and corrects errors by iteratively reviewing, remasking, and refining generated blocks. An illustrative example of this error correction cycle on an intermediate step of a trigonometric problem is provided in Appendix~\ref{app:qual_example}, specifically in Figure~\ref{fig:qual_example_cycle}.

\subsection{Quantitative Results}
We evaluated the accuracy (number of correctly solved problems) on the 127-question MATH subset. The results are summarized in Table \ref{tab:accuracy_results}.

\begin{table}[h!]
\centering
\caption{Accuracy on MATH 500 subset (127 questions). \(K\) is the R3 window size; \(N_S=5\) for R3 and BoN.}
\label{tab:accuracy_results}
\vskip 0.15in
\begin{center}
\begin{small}
\begin{sc}
\begin{tabular}{lcc}
\toprule
Method & Correct / Total & Accuracy (\%) \\
\midrule
Simple Diffusion (pass@1) & 37 / 127 & 29.13\% \\
R3 (\(K=4\)) & 42 / 127 & 33.07\% \\
R3 (\(K=5\)) & 44 / 127 & 34.65\% \\
\textbf{R3 (\(K=8\))} & \textbf{54 / 127} & \textbf{42.52\%} \\
Block-wise Best of N (BoN) & 61 / 127 & 48.03\% \\
\bottomrule
\end{tabular}
\end{sc}
\end{small}
\end{center}
\vskip -0.1in
\end{table}

\subsection{Discussion}
R3 with \(K=8\) substantially improves over the simple diffusion baseline, increasing accuracy from 29.13\% to 42.52\%, demonstrating the benefit of PRM-guided iterative refinement.

While block-wise BoN achieves the highest accuracy (48.03\%), it requires \(N_{\text{total}} \times N_S\) PRM calls (e.g., \(16 \times 5 = 80\) calls for a 16-block sequence). R3 is significantly more efficient: in the best case, it makes only \(\lceil N_{\text{total}} / K \rceil\) calls (e.g., 2 for \(K=8\)); in the worst case, \(2 \times \lceil N_{\text{total}} / K \rceil\) calls (4 for \(K=8\)). Thus, R3 achieves strong accuracy with far fewer PRM calls (2-4 vs. 80) by strategically targeting only problematic windows.

Larger window sizes improve performance: increasing \(K\) from 4 to 8 enhanced accuracy, suggesting that evaluating larger segments allows better contextual assessment and more impactful refinements. R3 offers a compelling balance between performance gains and computational efficiency.

\section{Conclusion}
\label{conclusion}

We introduced Review, Remask, Refine (R3), an inference-time framework that enhances masked diffusion models using Process Reward Models (PRMs). R3 iteratively generates text blocks, reviews them with a PRM, strategically remasks low-quality tokens, and refines these segments. Its windowed evaluation strategy enables targeted error correction without retraining.

Our experiments on MATH 500 showed R3 significantly improves accuracy over simple diffusion baselines, increasing from 29.13\% to 42.52\% with \(K=8\). R3 is also computationally efficient, requiring far fewer PRM calls than block-wise Best of N while achieving strong performance. Larger review windows proved beneficial, suggesting that more context enables better quality improvements.

R3 offers a promising approach for improving masked diffusion models by integrating process supervision at inference time. Future work could explore adaptive window sizes and sophisticated remasking strategies. A key research direction involves developing self-correction capabilities within the base model during post-training, potentially reducing reliance on external PRMs. Applying R3 to broader text generation tasks and diffusion architectures warrants further exploration.

\newpage
\bibliography{r3}
\bibliographystyle{icml2025}
\newpage

\appendix
\section{Qualitative Example of R3 Error Correction Cycle}
\label{app:qual_example}

Figure~\ref{fig:qual_example_cycle} illustrates the R3 error correction cycle. The framework identifies an initial error in a generated block, uses the PRM score to guide targeted remasking, and then refines the remasked segment to produce a corrected output.

\begin{figure}[h!]
\centering
\footnotesize
\begin{tabular}{p{.31\linewidth} p{.31\linewidth} p{.31\linewidth}}
\textbf{1. LLM Output – Before R3 Intervention (Initial Error)} &
\textbf{2. Conceptual State – After R3 Remasking (Based on Low PRM Score)} &
\textbf{3. LLM Output – After R3 Refinement} \\
\rule{\linewidth}{0.4pt} & \rule{\linewidth}{0.4pt} & \rule{\linewidth}{0.4pt} \\
Step 4: Determine the value of \(b\) \newline
The period \(T\) of the sine function \(y = a \sin(bx + c) + d\) is given by: \newline
\( T = \frac{2\pi}{b} \) \newline
From Step 3, we know \(T = \pi\). Therefore, we can set up the equation: \newline
\( \pi = \frac{2\pi}{b} \) \newline
Solving for \(b\): \newline
\( b = \frac{2\pi}{\pi} = \mathbf{3} \) % Incorrect part
&
\raggedright
\texttt{[MASK]}Step 4: Determine the value of ( b ) \newline
The period ($T$) of the sine function ($y = a \sin(bx + c) + d$) is given by: \newline
$T = \frac{2\pi}{b}$ \texttt{[MASK][MASK][MASK]} From \texttt{[MASK][MASK]} 3, \texttt{[MASK][MASK]} (\texttt{[MASK]} = \texttt{[MASK]}). Therefore \texttt{[MASK][MASK][MASK]} set \texttt{[MASK]} the equation: \newline
$\pi = \frac{2\pi}{b}$ \texttt{[MASK][MASK][MASK]} \newline
\texttt{[MASK]} \newline
Solving for \texttt{[MASK]} b \texttt{[MASK][MASK][MASK]}
&
Step 4: Determine the value of \(b\) \newline
The period \(T\) of the sine function \(y = a \sin(bx + c) + d\) is given by: \newline
\( T = \frac{2\pi}{b} \) \newline
From Step 3, we know \(T = \pi\). Therefore, we can set up the equation: \newline
\( \pi = \frac{2\pi}{b} \) \newline
Solving for \(b\): \newline
\( b = \frac{2\pi}{\pi} = \mathbf{2} \) % Corrected part
\\
\end{tabular}
\caption{Qualitative example of the R3 error correction cycle. (1) An initial error in calculation. (2) Conceptual representation of the block after the PRM ``Review" leads to targeted ``Remasking". (3) The ``Refine" stage produces the corrected calculation.}
\label{fig:qual_example_cycle} % Note: The label remains the same as the original figure
\end{figure}

\section{PRM Score to Remasking Probability Calculation}
\label{app:remasking_prob}
We detail the conversion of a PRM score \(S_b \in [0,1]\) for a block \(x_b\) into a remasking probability \(P_R(S_b)\). This probability determines the likelihood and extent of remasking for that block if its window triggers refinement.
\begin{enumerate}
    \item An intermediate quality value \(q_b\) is first computed from the PRM score \(S_b\) using an exponential decay function: \(q_b = \exp(-\alpha_B \cdot S_b)\). The hyperparameter \(\alpha_B\) (backmasking alpha, e.g., 10.0 in our experiments) controls the steepness of this transformation; a higher \(\alpha_B\) makes the \(q_b\) value more sensitive to small differences in \(S_b\), especially penalizing higher scores more rapidly.
    \item These \(q_b\) values for all blocks within the current scoring window \(W\) are then normalized to a probability range \([p_{\text{min}}, 1]\). Let \(\{q_{b'}\}_{b' \in W}\) be the set of these intermediate values for the window. The final remasking probability for block \(x_b\) is:
    \[ P_R(S_b) = p_{\text{min}} + (1 - p_{\text{min}}) \cdot \frac{q_b - \min_{b' \in W} q_{b'}}{\max_{b' \in W} q_{b'} - \min_{b' \in W} q_{b'} + \epsilon} \]
    where \(p_{\text{min}}\) is a minimum remasking probability (e.g., 0.01) to ensure even high-scoring blocks have a small chance of being re-evaluated, and \(\epsilon\) is a small constant for numerical stability. This formula effectively maps lower PRM scores (which result in higher \(q_b\)) to higher remasking probabilities.
\end{enumerate}

\end{document}